\documentclass[letterpaper, 10 pt, conference]{ieeeconf}  

\IEEEoverridecommandlockouts                              

\usepackage{graphicx}
\usepackage{tabularx}
\usepackage{amsmath}
\usepackage{amssymb}
\usepackage{cite}
\usepackage{booktabs}
\makeatletter
\let\NAT@parse\undefined
\makeatother
\usepackage[pdftex,
        colorlinks=true,
        bookmarks=true,
        citecolor=red,
        linkcolor=blue,
        pagebackref=false,
        pdfstartview=Fit,
        breaklinks=true
            ]{hyperref}
\usepackage{cleveref}
\usepackage{algorithm, algpseudocode}

\title{\LARGE \bf
DynaMIC: Dynamic Multimodal In-Context Learning Enabled Embodied Robot Counterfactual Resistance Ability
}

\author{Tianqiang Yan$^{1}$, Ziqiao Lin$^{2}$, Sicheng Wang$^{3}$, Tianwei Zhang$^{3,*}$ and Zhenglong Sun$^{2,3,*}$
\thanks{$^{1}$Faculty of Information Technology, Monash University, Melbourne, Australia}
\thanks{$^{2}$School of Science and Engineering, the Chinese University of Hong Kong-Shenzhen, Shenzhen, China}%
\thanks{$^{3}$Shenzhen Institute of Artificial Intelligence and Robotics for Society, the Chinese University of Hong Kong-Shenzhen, Shenzhen, China}%
\thanks{$^{*}$Corresponding author:\{zhangtianwei, zhenglongsun\}@cuhk.edu.cn}%
\thanks{This work was supported by the National Natural Science Foundation of China (Grant No. 62306185); 
the Guangdong Basic and Applied Basic Research Foundation (Grant No. 2023B1515020089) and the Shenzhen Science and Technology Program (Grant No. JSGGKQTD20221101115656029 and KJZD20230923113801004).}
}

\begin{document}

\maketitle
\thispagestyle{empty}
\pagestyle{empty}
\begin{abstract}
The emergence of large pre-trained models based on natural language has breathed new life into robotics development. Extensive research has integrated large models with robots, utilizing the powerful semantic understanding and generation capabilities of large models to facilitate robot control through natural language instructions gradually. However, we found that robots that strictly adhere to human instructions, especially those containing misleading information, may encounter errors during task execution, potentially leading to safety hazards. This resembles the concept of counterfactuals in natural language processing (NLP), which has not yet attracted much attention in robotic research. In an effort to highlight this issue for future studies, this paper introduced directive counterfactuals (DCFs) arising from misleading human directives. We present DynaMIC, a framework for generating robot task flows to identify DCFs and relay feedback to humans proactively. This capability can help robots be sensitive to potential DCFs within a task, thus enhancing the reliability of the execution process. We conducted semantic-level experiments and ablation studies, showcasing the effectiveness of this framework. 

\end{abstract}

\begin{figure}[!h]
  \centering
  \includegraphics[width=0.47\textwidth]{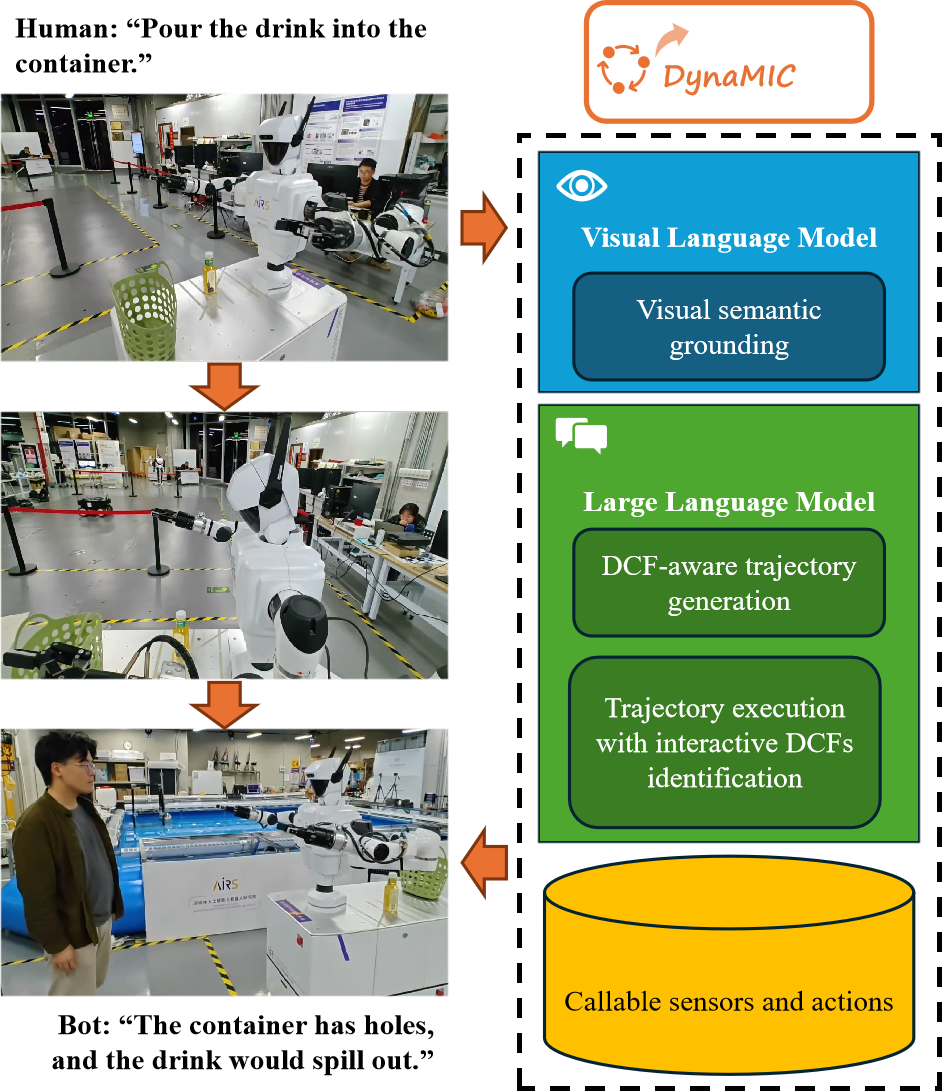}
  \caption{We proposed DynaMIC, a Dynamic Multimodal In-Context learning framework for dynamic DCF-aware robot task planning with interactive natural language feedback. DCFs refer to condition threatening the reliability of large model-driven robots under biased human instructions. \textbf{Left} illustrations of a robot identifying a DCF, and providing the human with feedback about the condition. \textbf{Right} a brief architecture of DynaMIC.}
  \label{main_idea}
  \vspace{-0.5cm}
\end{figure}

\section{INTRODUCTION}
Large pretrained models based on natural language, e.g. large language models (LLMs) and visual language models (VLMs)~\cite{brown2020language, achiam2023gpt, yang2023dawn}, etc., have pioneered a paradigm shift for the field of robot learning, leading traditional robots towards the ultimate goals as embodied AIs and artificial general intelligence (AGI)~\cite{xu2024survey, duan2022survey}. An increasing number of studies have leveraged the powerful semantic understanding, reasoning, and generation capabilities of these models to achieve instruction-based and autonomous task planning for robots facing universal application scenarios~\cite{singh2023progprompt, jiang2022vima, wake2023gpt}. However, in the real-world human-robot interaction, while previous work mainly focused on enabling robots to understand and execute human instructions, one of the potential issues could still bring significant challenges to the interaction reliability. In this article, this issue is defined as \textit{ direct counterfactuals (DCFs)}, of which the concept ``counterfactual'' originates from~\cite{calderon2022docogen} to describe knowledge that contrasts with facts for the robustness analysis of NLP models~\cite{lee2022deduplicating, lee2022factuality, li2020slot, mesgar2021improving}. It describes the situation where the robot might be misled by human instructions that contain such counterfactual information.

We provide a basic example to illustrate the scenarios in which DCFs occur. For example, the user commands the robot, ``I am thirsty, pass me the beverage on the table''. However, the beverage bottle on the table might be empty; passing an empty bottle to the person would not satisfy their intended purpose.  In this example, the person lacks direct access to the bottle, whereas only the robot can physically interact with it. In this circumstance, when the robot grasps the bottle and senses its weight, an information asymmetry arises between the person and the robot. This disparity in understanding the status of objects in the real-world environment could lead to a counterfactual scenario. It is expected that the robot can dynamically interact with the user notifying such counterfactualness, rather than blindly executing the user's commands.

Recent research has highlighted the importance of the reliability of large model-driven robots, particularly in contexts involving human-robot interaction and complex environmental structures~\cite{zhang2023building, sarch2023open, geng2023sage}. 
\cite{zhang2023building} advocated for the robot's information advantage towards its operational environment as a catalyst for improved human-robot collaboration, positing that a robot's more detailed understanding of its surroundings enables more efficient task execution.~\cite{sarch2023open} introduced a continually instructable framework that allowed robots to adjust their actions based on human commands and modify their movements accordingly in real time.~\cite{geng2023sage} delivered a large model-driven framework for robot's autonomous object manipulation based on human instructions, emphasizing the framework's ability to guide robots to self-adjust and re-plan tasks when encountering problems. However, their feedback and re-planning module refers to identifying failures and adjusting strategies during action execution if there is a mismatch between actual component movements and expectations. This mode of re-planning after encountering failures still poses reliability issues in scenarios with high trial-and-error costs. To the best of our knowledge, recent studies on large model-driven robots have not investigated much about such issues that may raise potential threats to the reliability of these robots. In addition, with few concerns about misleading human instructions, hardly any research has delved into the targeted solutions. Thus, in this paper, in addition to providing explanations for DCFs, we proposed a prompt-based in-context learning (ICL) framework~\cite{jin2022good, madotto2021few, rubin2022learning}, DynaMIC, for large models to assist robots in planning task flows~\cite{gu2021response, zhang2023large, vemprala2023chatgpt} where DCFs could occur, as shown in Fig~\ref{main_idea}.

DynaMIC comprises a locally distributed VLM (integrated with the robot), an embedded knowledge base storing information on a robot's callable sensors and actions, and a cloud-based LLM. The framework is designed to assist the robot by dynamically planning a counterfactual-insightful task flow based on the multimodal perception of the environment and the intended purpose of a human. The ICL mechanism operates under a structural prompt-based finite state machine (FSM)~\cite{yan2023refining, chen2023disco, tonmoy2024comprehensive, chen2023unleashing}. It effectively integrates all inputs and intermediate outputs of semantic information, identifying potential counterfactual situations within each operational state throughout the execution procedure, aiming to identify DCFs before the next action and provide timely feedback to humans.

In summary, our contributions can be outlined as follows:

\begin{itemize}
  \item We heuristically discuss the directive counterfactuals (DCFs), a concept drawing from recent LLM studies, to human-robot interaction.
  \item We presented DynaMIC, a LLM-based framework that aids dynamic task planning with DCF-awareness and is transferable to existing ICL methods. 
  \item Through real-world validation, we preliminarily demonstrate the efficacy of DynaMIC in enhancing human-centric robot operation. The results reflect the significance of introducing DCF-awareness, laying a foundation for further research.
\end{itemize}



\begin{figure*}[!htbp]
  \centering
  \includegraphics[width=\textwidth]{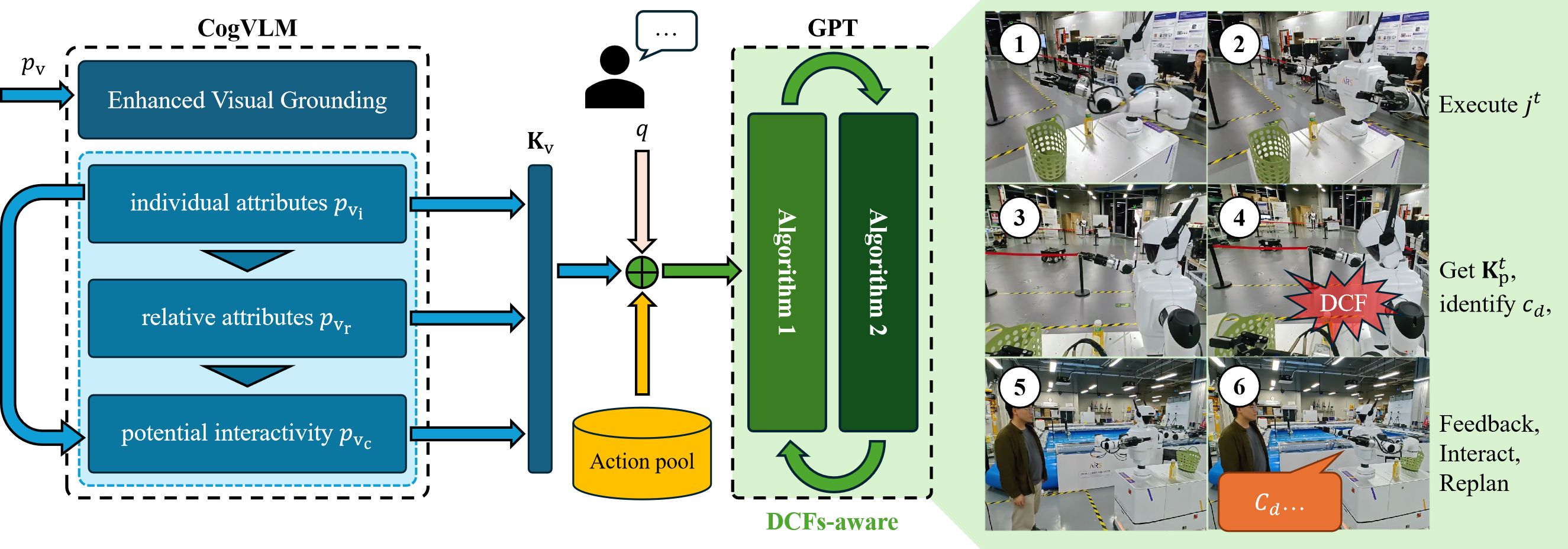}
  \caption{The workflow of DynaMIC, incorporating an example where a robot, during its task execution, identifies a DCF and gives feedback to the user about this information. For the \textbf{Algorithms \textbf{1}} and \textbf{2} utilized by the LLM, please refer to the corresponding pseudo-codes outlined in the article.}
  \label{architect}
\end{figure*}

\section{METHODOLOGY}
A \textit{directive counterfactual} refers to a scenario in which a robot is faced with a situation where part of the instruction conflicts with fact upon receiving and executing an instruction from a human. This discrepancy results in the robot's inability to fulfill the intended objective of the instruction, or worse, leads to unwanted consequences to the surroundings and human participants.

\subsection{Problem Definition}
Before involving the discussion of counterfactual, we begin with elaborating how an expected, non-counterfactual task outcome should satisfy, including 1) alignment with the human's instruction, and 2) not resulting in any negative consequences. 

Starting from the fundamental elements within a general scenario, where a user commands a LLM-driven robot $G$ to achieve $q$ on the foundation of his/her prior understanding $\hat{\mathbf{K}}_\mathrm{u}$, and the robot performs $j^0$ as the semantic-level task feasibility check of $q$ based on its knowledge base $\mathbf{K}_\mathrm{m}$, where the outcome of ${j^0}$ conditioned on $q$, $\mathbf{K}_\mathrm{u}$, and $\mathbf{K}_\mathrm{m}$, denoted as $G(j^0 \mid q, \mathbf{K}_\mathrm{u}, \mathbf{K}_\mathrm{m})$, is binary between ``1'' (feasible) and ``0'' (infeasible). The upper corner marks $\cdot^*$ in this article refer to \textit{state indicators} rather than exponents.

We formulate DCF in regards to the dynamicity of the robot's operations. The literal explanation of a DCF is that human instructor has a unilateral prior knowledge of the operational environment, whereas \textbf{the robot executing the task has extra degrees of exploratory freedom, and can dynamically retrieve environmental information to develop a more comprehensive knowledge base of the environment}. When the information discrepancy forms between the robot perception and the prior knowledge possessed by the human agent, conflicting with the robot's subsequent intention of accomplishing the original objectives, this circumstance is termed as a DCF, of which the simplest kind is a DCF detected right after $j^0$, e.g. an instruction that goes against common knowledge. To extend the simplest DCF to a dynamic setting, we denote the perceptual knowledge base at the robot's disposal as $\mathbf{K}_\mathrm{p}^t$, where $t$ signifies the $t$-th time step. $\mathbf{K}_\mathrm{p}$ is a cumulative collection of the perceived knowledge, where $\mathbf{K}_\mathrm{p}^t$ encompasses all independent and identically distributed (i.i.d.) multimodal sensory data accumulated from the initial moment through to the $t$-th time step. Specifically speaking, its expansion is contingent upon the four conditions: 1) the task $q$ issued by a human, 2) the base knowledge $\mathbf{K}_\mathrm{m}$ of the agent, 3) the prior state of the sensory data collection $\mathbf{K}_\mathrm{p}^{t-1}$, and 4) the previous trajectory $\mathbf{J}$ of the robot, represented as $\mathbf{J}^t$ (conditioned on $q$, $\mathbf{K}_\mathrm{m}$, $\mathbf{K}_\mathrm{p}^{t-1}$, and $\mathbf{J}^{t-1}$). $\mathbf{J}$ includes not only locomotive actions but also sensory acquisitions. 
A DCF detected during the dynamic execution phase is represented by the knowledge discrepancy between the time-varying $\mathbf{K}_\mathrm{p}$ and $\mathbf{K}_\mathrm{u}$. A DCF $c_\mathrm{d}$ exists if, at the $t$-th step,

\begin{equation*}
    \begin{split}
        D(\mathbf{K}_\mathrm{p}^t, \mathbf{K}_\mathrm{u}) &> 0 \\
        \text{s. t. } G(j^0 \mid q, \mathbf{K}_\mathrm{u}, &\mathbf{K}_\mathrm{m}, \mathbf{K}_\mathrm{p}^t) \to 0 \text{.}
    \end{split}
    \label{exp_pcf}
\end{equation*}

$D$ is the quantified discrepancy between $\mathbf{K}_\mathrm{p}^t$ and $\mathbf{K}_\mathrm{u}$. It can be measured by calculating the element-wise semantic similarities between entities from the two knowledge base~\cite{reimers2020making}. Combined with the result of initial feasibility check as a static component, we derived a unified form of a DCF-elicited situation as

\begin{equation*}
    \begin{split}
        D(\mathbf{K}_\mathrm{u}, \mathbf{K}_\mathrm{m}) + &\sum_t D(\mathbf{K}_\mathrm{p}^t, \mathbf{K}_\mathrm{u}) > 0 \\
        \text{s. t. } G(j^0 \mid q, \mathbf{K}_\mathrm{u}, \mathbf{K}_\mathrm{m}) &= 0 \text{ or } G(j^0 \mid q, \mathbf{K}_\mathrm{u}, \mathbf{K}_\mathrm{m}^t) \to 0
    \end{split}
    \label{exp_dcf}
\end{equation*}

\subsection{The DynaMIC Framework}
We unveiled the detailed architecture of DynaMIC (\textbf{Dyna}mic \textbf{M}ultimodal \textbf{I}n-\textbf{C}ontext learning) for DCF-aware dynamic and interactive task planning of LLM-driven robots, as shown in Fig~\ref{architect}.

\vspace{1mm}
\subsubsection{Enhanced Visual Grounding with Visual Language Model}
Recent advancements in instructional embodied AI incorporate visual understanding with natural language reasoning as a fundamental approach to bridge the gap between the physical world and LLMs, thanks to the rapid development of VLMs~\cite{duan2022survey}. DynaMIC employs CogVLM, the state-of-the-art 17B VLM proposed by~\cite{wang2023cogvlm}. CogVLM consists of a 10B visual-semantic alignment model and a 7B chat model that are jointly pre-trained on massive open-domain visual-language conversational datasets, making it capable of grounding language precisely in visual affordances. In order to fully leverage the visual-semantic ICL capacity of CogVLM, we proposed a three-fold prompt-guided enhanced visual grounding framework:

\begin{itemize}
    \item \textbf{Independent visual attributes retrieval $p_\mathrm{v_\mathrm{i}}$.} In this stage, the VLM is prompted to ground independent visual attributes of different objects given the current frame of view $p_\mathrm{v}$, encompassing 1) inherent appearance: types, colors, construction features, and 2) characteristics that may be reflected from the inherent appearance: materials, purposes. The output $p_\mathrm{v_\mathrm{i}}$ of this stage is a textual description.
    \item \textbf{Relative visual attributes retrieval $p_\mathrm{v_\mathrm{r}}$.} The relative visual attributes are grounded between the visible objects include 1) relative sizes, 2) relative positions, and 3) relative structural differences. The output of this stage $p_\mathrm{v_\mathrm{r}}$ is conditioned on $p_\mathrm{v_\mathrm{i}}$ and $p_\mathrm{v}$.
    \item \textbf{Inter-object interactivity $p_\mathrm{v_\mathrm{c}}$.} $p_\mathrm{v_\mathrm{c}}$ is predicted with $p_\mathrm{v}$ and all the precedent outputs of the VLM. It is obtained by querying the VLM how objects in $p_\mathrm{v}$ can possibly interact with each other, what are the potential limitations within the interactions, and what outcomes can be brought by these interactions.
\end{itemize}

All of these outputs are combined as the current state of the visual knowledge $\mathbf{K}_\mathrm{v}$ in natural language.

\vspace{1mm}
\subsubsection{Identification of Counterfactuals with Large Language Model}
DynaMIC is embodied with GPT-4 Turbo, the state-of-the-art LLM from openAI~\cite{openai2024gpt}, as the task coordinator of the robot. It generates DCF-aware task trajectories regarding human instructions under different scenarios, proactively delivers feedback to humans when identifying DCFs, and replans the subsequent movements that concern further instructions from humans. These are achieved in a two-stage manner.

\noindent \textbf{DCF-aware task trajectory generation} Algorithm~\ref{algor:1} formulates the prompt-based procedure of the DCF-aware task trajectory generation, where the overall logic is divided by two key steps:

\begin{itemize}
    \item aligning $\mathbf{K}_\mathrm{v}$ and the intention of human instruction $q$,
    \item generating and examining an initial trajectory $\mathbf{J}_0$.
\end{itemize} 

The alignment between $\mathbf{K}_\mathrm{v}$ and the intention of $q$ aims to verify that the information in the instruction aligns with the predicted characteristics of the entities in the field. If the alignment cannot be met, $c_\mathrm{d}$ is raised from the visual perspective, the robot should provide a feedback to the human and wait for a revised $q'$. Until a $q'$ satisfies $\mathbf{K}_\mathrm{v}$, this procedure iterates.

\begin{figure*}[t]
  \centering
  \includegraphics[width=\textwidth]{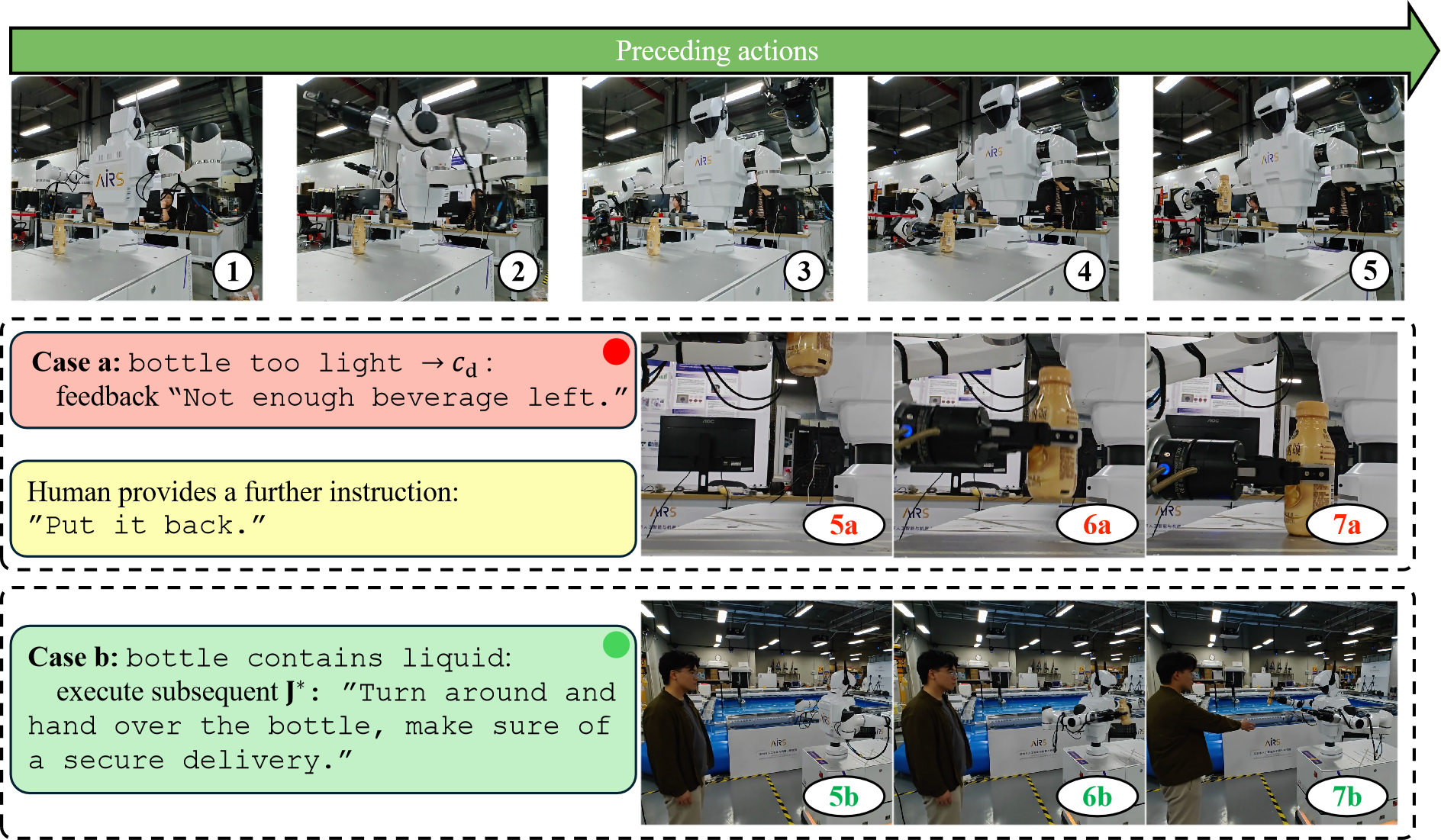}
  \caption{Demonstration of DynaMIC's helping the robot respond to different conditions regarding the liquid state of the bottle. 
  \textbf{Case a:} \texttt{bottle too light}, DynaMIC regarded the perception as a DCF to the user's instruction, thus telling the user ``\texttt{Not enough beverage left}''. The user provided a further instruction ``\texttt{Put it back}'', after which DynaMIC replanned the subsequent actions and directed the robot to put down the bottle. \\
  \textbf{Case b:} \texttt{bottle contains liquid}. In this case, the perception aligned with the user’s intention, thus the robot was guided to hand the bottle over.}
  \label{demo}
  \vspace{-2mm}
\end{figure*}

As soon as the intention of $q$ aligns with $\mathbf{K}_\mathrm{v}$, LLM is prompted to generate an initial trajectory $\mathbf{J}_0$ given the human's intention and the robot's capacity $\mathbf{K}_\mathrm{b}$, of which $\mathbf{K}_\mathrm{b}$ comprises the textual description of sensory modules equipped with the robot and the available actions of the robot in a JSON format. In order to prevent the embodied AI from neglecting important steps that may assist the robot in identifying possible $c_\mathrm{d}$, $\mathbf{J}_0$ is examined step-wise before executed. The LLM reasons every generated action $j$ and, if any perceptual action $j'$ within $\mathbf{K}_b$ should have been executed right after $j$ for $c_\mathrm{d}$ identification while not, $j'$ would then be inserted. Finally, $\mathbf{J}^*$, a refined version of $\mathbf{J}_0$, is outputted for the task execution phase.

\begin{algorithm}[!htb]
\caption{DCF-aware task trajectory generation with LLM}
\begin{algorithmic}[1]
\State \textbf{Inputs:} Human instruction $q$, grounded visual knowledge $\mathbf{K}_\mathrm{v}$, action pool $\mathbf{K}_\mathrm{b}$
\State \textbf{Output:} DCF-aware task trajectory $\mathbf{J}^*$
\State \textbf{Prompt templates:} Intention analysis request $r_\mathrm{i}$, visual-intention alignment analysis request $r_\mathrm{a}$, trajectory generation request $r_\mathrm{t}$
\Procedure{Traj-PreGen}{$q, \mathbf{K}_\mathrm{v}, \mathbf{K}_\mathrm{b}$}
\While{True}
    \State prompt $x_1 \gets q$ and $r_\mathrm{i}$
    \State intention $\xi \gets G(x_1)$
    \State $x_2 \gets \xi$ and $\mathbf{K}_\mathrm{v}$
    \If{NOT $G(x_2) = $ True}
        \State \textbf{return} False $\to$ human
        \While{waiting for a revised $q'$}
            \If{LLM $\gets q'$ from human}
                \State $q \gets q'$
                \State \textbf{break}
            \Else
                \State \textbf{return} $q$ NOT executable $\to$ human
            \EndIf
        \EndWhile
    \Else
        \State \textbf{break}
    \EndIf
\EndWhile
\State $x_3 \gets \xi$, $\mathbf{K}_\mathrm{v}$, and $\mathbf{K}_\mathrm{b}$, $\mathbf{J}_0 \gets G(x_3)$ 
\For{every $j$ in $\mathbf{J}_0$}
    \State \textbf{if} $\exists j' \in \mathbf{K}_\mathrm{b}$ \& $\notin \mathbf{J}_0$ s. t. $j'$ helps identify $c_\mathrm{p}$
    \State \textbf{then}
        \State \hspace{1.5em} $\mathbf{J}_0$ is updated by inserting $j'$ after $j$
    \State \textbf{end if}
\EndFor
$\mathbf{J}^* \gets \mathbf{J}_0$
\State \textbf{return} $\mathbf{J}^*$
\EndProcedure
\end{algorithmic}
\label{algor:1}
\end{algorithm}

\noindent \textbf{Interactive DCFs identification and Task Replanning} An edge-level natural language alignment model, MiniLM~\cite{wang2021minilmv2}, is chosen for aligning every actions in $\mathbf{J}^*$ with low-level robot action planner $P$, of which the corresponding control command lines are also indexed in the action pool $\mathbf{K}_\mathrm{b}$. Since the purpose of this thesis is not about accurate end-effector trajectory planning, therefore the scenario setup is simplified, where complicated end-effector trajectory is not involved. Besides, in order to interpret the raw data collected by edge sensors other than the camera, i.e. haptic sensor, a rule-based data transcriber $T$ is introduced to describe the fluctuation (e.g. ascending, plateau, descending, oscillating, etc.) characteristics of data values regarding statistical calculations in pre-defined natural language templates. Algorithm~\ref{algor:2} executes the trajectory returned from Algorithm~\ref{algor:1}, dynamically identifies DCFs, interacts with the human agent, and cooperates with Algorithm~\ref{algor:1} to replan subsequent trajectory when it comes to a DCF that could lead to the failure of a task.

\begin{algorithm}[!h]
\caption{Interactive DCFs identification and task replanning}
\begin{algorithmic}[1]
\State \textbf{Input:} human instruction $q$, trajectory $\mathbf{J}^*$, multimodal perceptual knowledge $\mathbf{K}_\mathrm{p}$
\Procedure{Identify\&Replan}{$q, \mathbf{J}^*, \mathbf{K}_\mathrm{p}$}
\For{$j$ in $\mathbf{J}^*$}
    \State call $P(\mathrm{step})$
    \State \textbf{case} $j$ is a movement: \textbf{continue}
    \State \textbf{case} $j$ is a perception:
    \State \hspace{1.5em} call $T(\text{sensory data}) \to \mathbf{K}_\mathrm{p}$
    \State \hspace{1.5em} LLM $\gets \mathbf{K}_\mathrm{p}$ and identifies $c_\mathrm{d}$
    \
    \If{$c_\mathrm{p} = $ True}
        \State \textbf{try}
        \State \hspace{1.5em} call Algorithm~\ref{algor:1} to revise $\mathbf{J}^*$
        \State \hspace{1.5em} $\mathbf{J}^* \gets \mathbf{J}^{*'}$ from Algorithm~\ref{algor:1}
        \State \hspace{1.5em} \textbf{continue}
        \State \textbf{except} $\mathbf{J}^*$ NOT refinable given $q$ and $\mathbf{K}_\mathrm{b}$ 
        \State \hspace{1.5em} \textbf{return} $q$ NOT executable $\to$ human
        \State \hspace{1.5em} call Algorithm~\ref{algor:1} and await new $q'$
    \EndIf
\EndFor
\EndProcedure
\end{algorithmic}
\label{algor:2}
\end{algorithm}

\section{EXPERIMENTS}

\subsection{Scene Settings}
\begin{figure}[h]
  \centering
  \includegraphics[width=\linewidth]{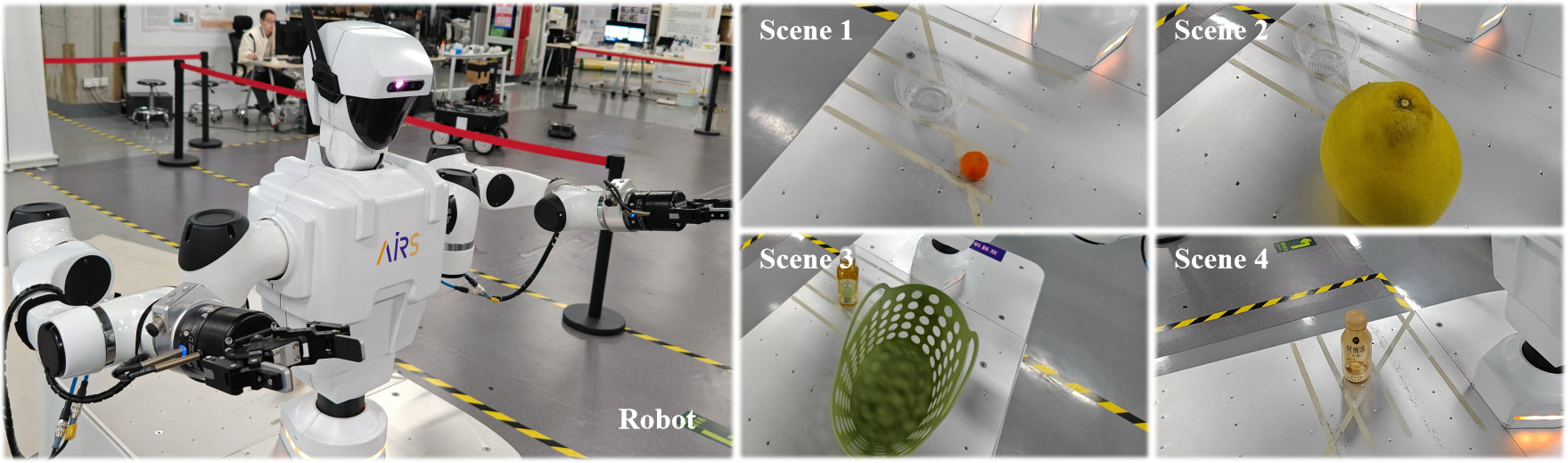}
  \caption{Illustrations of the representative scenes included in our experiments. Scene 1 shows an orange fruit and an empty plastic bowl on the table, under which the instruction is ``put the fruit into the bowl''; Scene 2 shows a pomelo and an empty plastic bowl on the table, under which the instruction is ``put the fruit into the bowl''; Scene 3 shows a bottle of beverage and a basket with holes on the table, under which the instruction is ``pour the drink into the container''; Scene 4 shows a bottle of beverage with an opaque body, under which the instruction is ``I'm thirsty. Pass me the beverage''.}
  \label{scenarios}
  \vspace{-2mm}
\end{figure}

We set up 10 life-based experimental scenes with corresponding user instructions to testify and showcase the design objectives of DynaMIC. The illustrations of four representative scenes are shown in Fig~\ref{scenarios}. 

Scenes 1 to 3 primarily focus on identifying visual DCFs. Among them, scene 1 does not have a predetermined DCF; in scene 2, the preset DCF is that the fruit on the table is much larger than the bowl, thus preventing the fruit from being put into the bowl; scene 3's DCF involves the container on the table that is a basket with holes, therefore it cannot contain liquids. scene 4 shifts focus to the identification of non-visual DCFs. In this scene, the beverage level inside the bottle is randomly determined to be either sufficient or nearly depleted. With sufficient beverage, based on the given instruction ``I'm thirsty. Pass me the beverage on the table'', there is no DCF present; nonetheless, in the case where the beverage level is nearly depleted, a DCF arises because the state of ``an empty bottle'' contradicts the intention of the instruction ``I want to drink the beverage''. The rest of the scenes are described in ~\ref{scenes}.

\subsection{Performance Demonstration and Evaluation}
We evaluated DynaMIC's outputs across all scenes, ensuring result stability by conducting five iterative tests for each scene. Out of 10 scenarios, a total of 50 tests were conducted, of which 44 resulted in the expected output and feedback. In scene 4, as an example, with the exception of one round of test where DynaMIC's task trajectory included a movement beyond the robot's sensory and action capabilities, it successfully activated the perception of bottle weight in the remaining four rounds. In Fig~\ref{demo}, we showcase how DynaMIC directs the robot's subsequent actions in scene 4 for two distinct situations: one where the bottle retains a sufficient amount of beverage, the other where it is nearly empty.

\begin{table}[h]
\caption{Result of Ablating Visual Semantic Grounding Outputs}
\label{abla_vlm}
\centering
\begin{tabularx}{0.48\textwidth}{c|cc|cc|cc}
\toprule
Ablated Output & \multicolumn{2}{c|}{$p_{\mathrm{v}_\mathrm{i}}$} & \multicolumn{2}{c|}{$p_{\mathrm{v}_\mathrm{r}}$} & \multicolumn{2}{c}{$p_{\mathrm{v}_\mathrm{c}}$} \\
\midrule
Scene ID & 2 & \textbf{3} & \textbf{2} & 3 & 2 & 3 \\
\midrule
DCFs identified (X/5) & 5 & \textbf{2} & \textbf{0} & 5 & 5 & 5 \\

DCFs Identified \textit{No Ablation} (X/5) & 5 & \textbf{5} & \textbf{5} & 5 & 5 & 5 \\
\bottomrule 
\end{tabularx}
\end{table}

\begin{table}[h]
\caption{Result of Ablating DCF-aware Trajectory Refinement}
\label{abla_llm}
\centering
\begin{tabularx}{0.47\textwidth}{c|c}
\toprule
Ablated module & Refinement phase in Algorithm~\ref{algor:1} \\
\midrule
Scene ID & \textbf{4} \\
\midrule
DCFs identified (X/5) & \textbf{3} \\
DCFs identified \textit{no ablation} (X/5) & \textbf{5} \\
\bottomrule 
\end{tabularx}
\end{table}

\begin{table*}[t]
\caption{Robot’s performance on life-based experimental scenes}
\label{scenes}
\centering
\begin{tabularx}{\textwidth}{X|X|X|X}
\toprule
\textbf{Scene} & \textbf{Command} & \textbf{Robot's performance without DynaMIC} & \textbf{Robot's performance with DynaMIC} \\
\midrule
A room with the lights already turned off & ``Turn off the lights in the room.'' & Attempt to turn off the lights, even though they are already off & Inform the user that the lights are already off \\
\midrule
A closed door with a ``Do Not Disturb'' sign & ``Please enter the room and bring me the documents on the desk.'' & Enter the room despite the ``Do Not Disturb'' sign & Inform the user about the ``Do Not Disturb'' sign \\
\midrule
An opaque and sealed snack container on the shelf & ``I'm hungry, pass me the snack from the shelf.'' & Hands the snack container to the user & Assess the container's weight to check if there are snacks inside and inform the user if it's empty \\
\midrule
A closed shoe box on the shelf & ``I need my shoes from the box.'' & Hands the shoe box to the user & Assess the shoe box's weight to check if there are shoes inside and inform the user if it's empty \\
\midrule
A covered container of soup on the stove & ``Serve me some hot soup from the container.'' & Serve the soup without checking its temperature & Check the soup's temperature to ensure it's hot and inform the user if it’s not \\
\midrule
A pot of coffee on the table & ``Pour me some hot coffee from the pot.'' & Pour the coffee without verifying its temperature & Check the coffee's temperature to ensure it's hot and inform the user if it's not \\
\bottomrule
\end{tabularx}
\end{table*}

\noindent \textbf{Ablating visual semantic grounding outputs} We conducted ablation studies by removing each of the three visually grounded outputs, i.e. $p_{\mathrm{v}_\mathrm{i}}$, $p_{\mathrm{v}_\mathrm{r}}$, and $p_{\mathrm{v}_\mathrm{c}}$, from the VLM of DynaMIC at a time, under scenes 2 and 3. The consistency of the result was again guaranteed by iterating each ablation study five rounds. We reported the result in Table~\ref{abla_vlm}. For scene 2, removing $p_{\mathrm{v}_\mathrm{r}}$ resulted in the lack of relative perception of the size difference between the fruit and the bowl, hence suggesting no misalignment between the user's intention and the visual information. Regarding scene 3, the ignorance of $p_{\mathrm{v}_\mathrm{i}}$ concealed the fact that the container is a basket with holes from being noticed, leading to the omission of the DCF. The ablation of $p_{\mathrm{v}_\mathrm{r}}$ had a devastating impact on the DynaMIC's performance under scene 2, rendering an all-missed result. The ablation of $p_{\mathrm{v}_\mathrm{i}}$ also degraded the functionality of DynaMIC under scene 3. Though it did not cause DynaMIC to miss the DCF in every round of test, the two ``successful'' identifications actually stemmed from the accidental appearance of independent attributes within $p_{\mathrm{v}_\mathrm{r}}$ according to the tracebacks of the outputs, implying that such instances are purely by chance.

\vspace{1mm}
\noindent \textbf{Ablating DCF-aware trajectory refinement} This ablation phase was conducted under scene 4. It was observed that DynaMIC could occassionally demonstrate insight into potential DCFs within the pre-generated $\mathbf{J}_0$ even without the static refinement phase as stated in Algorithm~\ref{algor:1}. This thesis hypothesized such situations originates from the inherent corpus of the LLM containing certain prior knowledge, which includes human habitual validations for some counterfactual situations. This type of prior knowledge can be injected into the trajectories generated by the LLM in some cases. Therefore, the purpose of this ablation study is to examine the universality of this innate insight of an LLM and whether this refinement phase indeed significantly enhances LLM's insight into DCFs. 

The result displayed in Table~\ref{abla_llm} marks a decline in the identification of DCFs upon the removal of refinement phase in Algorithm~\ref{algor:1}. This result suggests that sole reliance on the LLM's prior knowledge against conterfactuals is still unreliable, and prompting the LLM to refine trajectories with DCF-aware actions before executing is significant for enhancing the identification of potential DCF conditions.


\section{CONCLUSION}

In this paper, we point out that robots blindly following human instructions are highly susceptible to counterfactuals induced by human-elicited biases. Inspired by the definitions of counterfactuals in LLM studies, we extend it by presenting the concept of DCFs, with a specific concentration on its dynamic and multimodal attributes. By embedding a VLM alongside a LLM in a humanoid robot, we showcase how our DynaMIC framework effectively identifies DCFs using multimodal sensory inputs. This allows for the detection and prevention of irreversible situations in fluctuating operational environments. The robust semantic reasoning abilities of these LLMs enable them to adapt to diverse scenarios, anticipate potential DCFs, swiftly adjust their strategies, and actively engage with humans.”
Overall, this paper offers a heuristic perspective for future studies on practical applications of human-robot interactions.




\bibliographystyle{IEEEtran}
\bibliography{reference}

\end{document}